\pdfoutput=1

\documentclass[11pt]{article}

\usepackage{acl}

\usepackage{enumitem}
\setlist[itemize]{leftmargin=*}

\usepackage[colorinlistoftodos,prependcaption,textsize=tiny]{todonotes}

\usepackage{comment}
\usepackage{array}
\usepackage{fancybox}
\usepackage{booktabs}
\newcolumntype{P}[1]{>{\centering\arraybackslash}p{#1}}
\newcommand{\ourdataset}{ChattyChef}
\newcommand{\fakemultirow}[1]{\multirow{2}{*}{\centering #1}}
\usepackage{xcolor,soul}

\usepackage{array}
\newcolumntype{L}[1]{>{\raggedright\let\newline\\\arraybackslash\hspace{0pt}}m{#1}}
\newcolumntype{C}[1]{>{\centering\let\newline\\\arraybackslash\hspace{0pt}}m{#1}}
\newcolumntype{R}[1]{>{\raggedleft\let\newline\\\arraybackslash\hspace{0pt}}m{#1}}

\definecolor{mygreen}{RGB}{50, 205, 50}
\definecolor{mygreen2}{RGB}{199, 250, 194}
\definecolor{myred}{RGB}{235, 135, 120}
\definecolor{myorange}{RGB}{242, 183, 112}
\definecolor{myorange2}{RGB}{240, 152, 55}
\definecolor{myblue}{RGB}{139, 176, 208}
\definecolor{myblue2}{RGB}{255,153,255}
\definecolor{weiblue}{RGB}{0, 77, 128}
\definecolor{weired}{RGB}{181, 23, 0}

\usepackage{times}
\usepackage{latexsym}
\usepackage{graphicx}
\usepackage[linesnumbered, ruled,vlined]{algorithm2e}
\usepackage{amsmath}
\usepackage{multirow}
\usepackage[T1]{fontenc}

\usepackage[utf8]{inputenc}

\usepackage{microtype}

\DeclareMathOperator*{\argmax}{arg\,max}

\title{Improved Instruction Ordering in Recipe-Grounded Conversation}

\author{Duong Minh Le, Ruohao Guo, Wei Xu, Alan Ritter \\
  Georgia Institute of Technology \\
  {\small \texttt{\{dminh6, rguo48\}@gatech.edu; \{wei.xu, alan.ritter\}@cc.gatech.edu}}}

\SetKwInput{KwInput}{Input}
\SetKwInput{KwOutput}{Output}

\begin{document}
\maketitle

\cornersize{1}

\begin{abstract}
In this paper, we study the task of instructional dialogue and focus on the cooking domain. Analyzing the generated output of the GPT-J model, we reveal that the primary challenge for a recipe-grounded dialog system is how to provide the instructions in the correct order. We hypothesize that this is due to the model's lack of understanding of user intent and inability to track the instruction state (i.e., which step was last instructed). Therefore, we propose to explore two auxiliary subtasks, namely User Intent Detection and Instruction State Tracking, to support Response Generation with improved instruction grounding. Experimenting with our newly collected dataset, \ourdataset, shows that incorporating user intent and instruction state information helps the response generation model mitigate the incorrect order issue. Furthermore, to investigate whether ChatGPT has completely solved this task, we analyze its outputs and find that it also makes mistakes (10.7\% of the responses), about half of which are out-of-order instructions. We will release \ourdataset~to facilitate further research in this area at: \url{https://github.com/octaviaguo/ChattyChef}. 
\end{abstract}

\section{Introduction}
Historically, work on conversational agents has mostly fallen into one of two categories: open-domain chatbots \citep{ritter2011data,li2016deep,thoppilan2022lamda,shuster2022blenderbot} or goal-directed dialogue systems within narrow domains \citep{williams2016dialog,eric2020multiwoz}.  However, recent advances in large language models have paved the way for the exploration of dialog agents that can engage in conversations with users to accomplish open-ended objectives, such as learning about a new topic \citep{dinan2019wizard,choi-etal-2018-quac,reddy2019coqa}, interpreting bureaucratic policies to answer questions \citep{saeidi2018interpretation}, or negotiating within strategy games \citep{lewis2017deal,meta2022human}.



\begin{table}[t]
\small
\centering
\begin{tabular}{ll|c}
    \toprule
    \multicolumn{2}{c}{\textbf{Correct Response}} & \multicolumn{1}{c}{49.6\%} \\ 
    \midrule
    \multicolumn{1}{c}{\multirow{5}{*}{\parbox{0.33\linewidth}{\vspace{.2cm}\textbf{Incorrect Response}}}} & \multicolumn{1}{c}{Wrong order} & \multicolumn{1}{c}{22.9\%} \\ \cmidrule{2-3} 
    \multicolumn{1}{l}{} & \multicolumn{1}{c}{Irrelevant response} & \multicolumn{1}{c}{10.7\%}  \\ \cmidrule{2-3} 
    \multicolumn{1}{l}{} & \multicolumn{1}{c}{Lack of information} & \multicolumn{1}{c}{8.4\%}  \\ \cmidrule{2-3} 
    \multicolumn{1}{l}{} & \multicolumn{1}{c}{Wrong information} & \multicolumn{1}{c}{8.4\%}  \\ 
    \bottomrule
\end{tabular}
    \caption{Manual analysis of 10 recipe-grounded conversations (131 responses in total) generated by a fine-tuned GPT-J model on the test portion of our new dataset \includegraphics[width=1em]{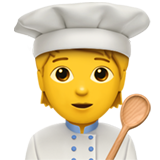} \ourdataset. The incorrect responses are classified into four error types (examples in Figure \ref{fig:example}) with out-of-order instructions being the most common.}
    
    \label{tab:error_stats}
\end{table}

\begin{figure*}[t]
    \centering
    \includegraphics[width=\textwidth]{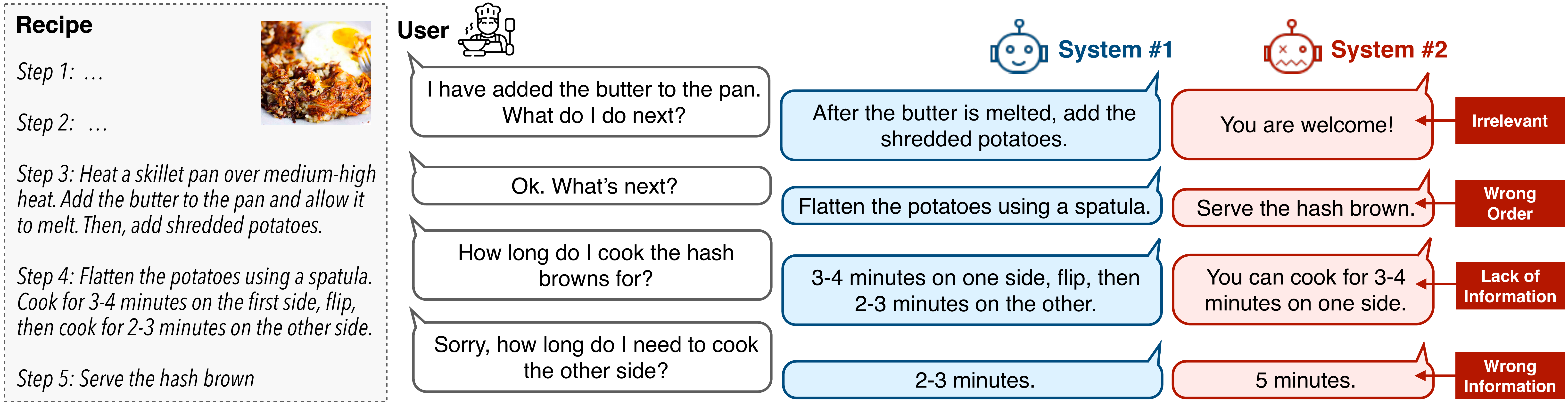}
    \caption{A conversation snippet of the cooking instructional dialogue task with \textbf{\color{weiblue}good} and \textbf{\color{weired}bad} system responses and the corresponding error type of each bad response.}
    \label{fig:example}
\end{figure*}

In this paper, we explore the task of {\em Recipe-Grounded Conversation}, where the dialogue agent is expected to converse with a user to walk him/her through the cooking procedure of a recipe, while answering any questions that might arise along the way (see examples in Figure \ref{fig:example}).  
Although many types of dialogue tasks have been proposed and explored, very little prior work has focused on providing instructions to a user to complete a task.  In contrast to other dialogue tasks, such as document-grounded conversation \cite{dinan2019wizard}, accurately tracking the conversation state is more crucial in recipe-grounded dialogue.  This is because the agent needs to know which step in the recipe the user is currently working on in order to answer questions, such as: {\em what is the next step?}

To investigate what challenges may arise in recipe-grounded conversation, we have collected a dataset by crowdsourcing (see \S \ref{sec:data}). As an initial baseline model, we used this data to fine-tune GPT-J following a similar protocol to \citet{peng2022godel}. Specifically, the conversation history, the grounded recipe, and the gold system response are concatenated as a long input sequence to the model (see \S \ref{sec:response_generation} for details). We show that fine-tuning GPT-J on a few recipe-grounded conversations works surprisingly well, however, the model makes a significant number of mistakes based on  our manual inspection over 10 conversations (131 generated responses) of the fine-tuned model (Table \ref{tab:error_stats}). Examples of each type of error are presented in Figure \ref{fig:example}. Notably, the most prevalent type of errors is presenting information from the recipe to the user in the wrong order.  
We thus focus on tackling this most common error in our work. We hypothesize two potential causes: (1) GPT-J struggles to understand the user's intent, and (2) GPT-J has difficulty tracking the current state throughout the conversation.  Both are crucial in many scenarios, for example, when the user asks for more information about the current instruction, the system should not skip ahead to a later step of the recipe.  Based on these hypotheses, we experiment with two supplemental tasks to improve instruction ordering: User Intent Detection (\S \ref{sec:methods_intent}) and Instruction State Tracking (\S \ref{sec:state_tracking}).

The goal of {\em Intent Detection} is to classify the user's current intent within a fixed set of possibilities (e.g., ask for the next instruction or ask for details about ingredients).  Because the set of intents varies across domains,\footnote{For example, we would need a different set of intents to model instructions in Windows help documents \citep{branavan2010reading}.} we take a few-shot transfer learning approach to leverage existing dialogue intent datasets, such as MultiWOZ 2.2 \citep{budzianowski-etal-2018-multiwoz} and Schema Guided Dialogue \citep{rastogi2020towards}.  We show that incorporating natural language descriptions of intents \citep{zhao2022description} can enable more effective transfer.  For example, F1 score for detecting 19 different user intents in \ourdataset~increases from 32.0 to 65.1 when transferring from MultiWOZ (\S \ref{sec:experiments_intent}). In addition to Intent Detection, we also explore a simple yet effective method for {\em Instruction State Tracking}. State tracking aims to identify which recipe step the user is currently working on. We show that based on unigram F1 overlap, despite the approach's simplicity, we are able to identify the most relevant recipe step at each turn of the conversation with nearly 80\% accuracy.





The information from these two subtasks is then used to support {\em Response Generation} to improve instruction ordering (\S \ref{sec:response_generation}).  Specifically, instead of feeding the whole recipe into the generation model, we leverage the instruction state to select only the most relevant knowledge. 
To incorporate user intents, we enrich the input prompt to the model with natural language descriptions of the predicted intent. Experiments show that even though intent and instruction state predictions are not perfect, including this information in the Response Generation model helps mitigate the wrong-order issue. We release \includegraphics[width=1em]{images/cook_1f9d1-200d-1f373.png} \ourdataset, a new dataset of cooking dialogues, to support future work on instruction-grounded conversational agents.

\section{Dataset Construction}
\label{sec:data}

To collect a corpus of recipe-grounded conversations for fine-tuning and evaluating models, we first obtain WikiHow\footnote{\url{https://www.wikihow.com/Main-Page}} articles under the Recipes category from the data compiled by \citet{zhangetal2020reasoning}.   
We control the qualities of the recipes by only selecting articles that have a helpful vote rating greater than 75\% and have at least $5$ votes. 
We retain the images from the recipes in our dataset, but experiments in this paper only make use of recipe texts. 
Moreover, in order to improve the conversation quality and avoid crowd workers quitting in the middle of a long conversation, we remove recipes with more than $8$ steps. 

\subsection{Conversation Collection}
After getting the recipes, we then ask crowd workers to create conversation data by role-playing. There are two roles in each conversation of our dataset: an agent and a user. The agent is provided with a full recipe article and assumed to be an expert on cooking, while the user can only see the title of the recipe (i.e., the name of the dish). During the conversation, the agent needs to help the user complete a cooking task with the knowledge learned from the recipe and/or their common knowledge about cooking.
Different interfaces are used by crowd workers when they play the agent and the user (see Appendix \ref{appendix:collection_interface}). 

Crowd workers are instructed that conversations should be relevant to the provided recipe, and should also be natural and interesting. In our process, at each turn of a conversation, agents need to identify and highlight the relevant text span in the article, if present, before sending a message, but they are not allowed to answer by copying and pasting. Instead, the agent must rephrase the instruction in their own words. To facilitate more natural interactions, both workers can discuss guidelines and ask their partner to resend messages whenever they have confusion or disagreements, using a separate chat interface.\footnote{During data annotation, we formed teams of two workers and allowed them to communicate with each other via Slack.} 
Furthermore, in our preliminary study, we found that users tend to repeatedly send messages such as ``What is next?'' to simply urge the agent to move on without thinking or trying to learn the task. 
To encourage diverse conversations, we provide different dialog act prompts for annotators to choose from: ``teach a new step'', ``ask a question'', ``answer a question'', and ``other'' (see Appendix \ref{appendix:collection_interface}).
Diverse dialog acts such as asking and answering questions are encouraged with higher payments.


\subsection{Dataset Statistics}
\label{subsec:dataset_stats}
We summarize the statistics of our final dataset -- \ourdataset~-- and compare it with CookDial \cite{jiang2022cookdial} in Table \ref{data_stats}. Compared to Cookdial, even though \ourdataset~has fewer utterances per dialogue, our recipe steps are much longer, and each step includes multiple sentences or micro-steps (about 6.0 sentences per step on average). This feature sets our dataset aside from the CookDial, where nearly all recipe steps have only one short sentence or one single instruction. Having recipes with long, multi-sentence steps makes the conversation more diverse, giving crowd workers more freedom in choosing their own way of instructing, as some micro-steps can be done in parallel while others can be skipped. This attribute is important as it makes our dataset closer to a real-life setting, where the user will normally not strictly follow the order of steps in the recipe. As shown in Table \ref{tab:dataset_diversity}, the utterances from the agent in our dataset are much more diverse than CookDial and have instructions worded more differently from the grounded recipe.



\begin{table}[t]
    \centering
    \small
    \begin{tabular}{lcc}
         \toprule
          \textbf{Dataset} & \ourdataset & CookDial \\
         \midrule
         \textbf{\textit{Conversation Statistics}}  \\
         \#Dialogues & 267 & 260 \\
         \#Utterances per dialog & 26.0 & 35.0 \\
         \#Grounding recipes & 267 & 260 \\
         \midrule
         \textbf{\textit{Recipe Statistics}} \\
         \#Steps per recipe & 3.9 & 8.4 \\
         \#Tokens per recipe & 417.7 & 120.0 \\
         \#Sentences per step & 6.0 & 1.0 \\
         \#Tokens per recipe step & 70.1 & 14.4 \\
         \bottomrule
    \end{tabular}
    \caption{The statistics of our dataset and CookDial.}
    \label{data_stats}
\end{table}

\begin{table}[]
    \centering
    \setlength{\tabcolsep}{2pt}
    \resizebox{\linewidth}{!}{%
    \begin{tabular}{lcc}
    \toprule

    \multirow{2}{*}{\textbf{Dataset}}  & \textbf{Diversity (\%)}  & \textbf{N-gram overlap (\%)} \\ 

    & (1/2-gram) & (1/2/3/4/5-gram) \\
    \midrule
    CookDial &   18.7 / 38.1 & 44.6 / 24.4 / 15.7 / 10.6 / 7.4\\
    \ourdataset     &    26.0 / 53.6 & 30.2 / 12.0 /\enspace5.8 /\enspace3.4 /\enspace2.2 \\
    \bottomrule
    \end{tabular}}
    \caption{The statistic about the diversity of the agent's utterances in CookDial and \ourdataset. \textbf{Diversity} (dist-1/2): the number of unique unigrams/bigrams divided by the total number of all unigrams/bigrams in the conversations. \textbf{N-gram overlap}: the percentage of n-gram overlap between the recipes and agent responses.}
    \label{tab:dataset_diversity}
\end{table}

\begin{figure}[t]
    \centering
    \includegraphics[width=0.8\columnwidth]{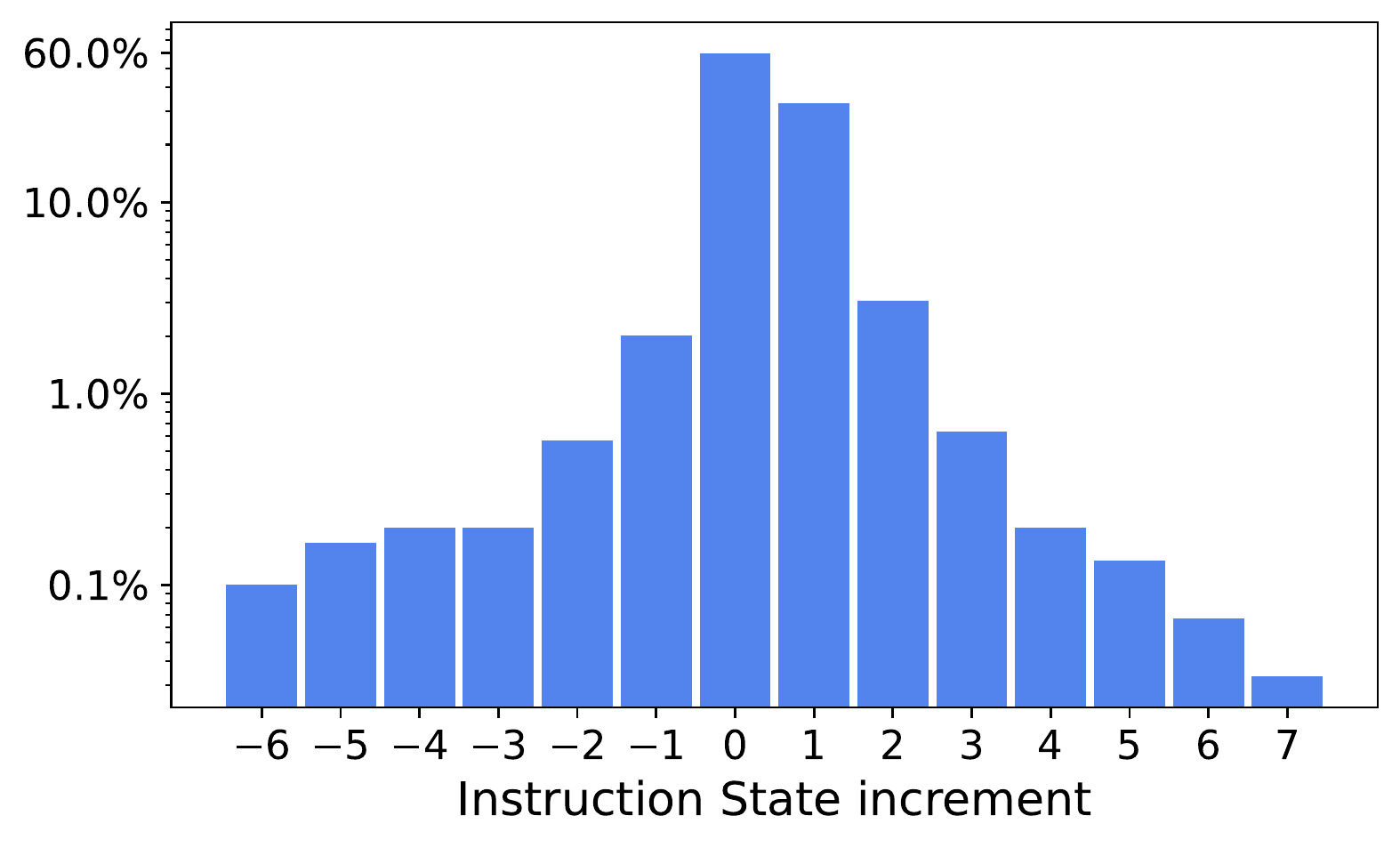}
    \caption{Statistic of the change of the instruction state between two consecutive agent utterances (the minus value means that the agent moves back and instructs on the past steps).}
    \label{fig:state_increment}
\end{figure}

\paragraph{Analysis of Instruction State Changes.}
In this work, we define the instruction state at a time step as the last recipe step which the agent instructed. We analyze  the change of instruction state between two consecutive agent utterances of our dataset in Figure~\ref{fig:state_increment}. Most of the time, the instruction would be either the same or the next recipe step. However, there are also cases when the agent needs to go back to previous steps (e.g., when the user requests to repeat an instruction) or go ahead (e.g., when the user wants to skip some steps) to provide instructions. In \ourdataset, the agent sometimes goes backward for as many as six steps, or forward seven steps. These observations have partially demonstrated the challenge of instructing multi-step procedures in the correct order, as there are many possibilities. Simply providing information from the recipe in a linear sequence is insufficient.

\section{User Intent Detection}
\label{sec:methods_intent}

In this section, we discuss the User Intent Detection subtask. Formally, the task is to predict a set of user intent indices $\mathcal{I}$ (as one utterance may contain multiple intents), given the $t^{\text{th}}$ user utterance $U^{usr}_t$ and the conversation history $\mathcal{H} = \{U^{sys}_1, U^{usr}_1, ..., U^{sys}_{t-1}\}$. We hypothesize that providing information about the user's intents may help the response generation model better provide information to the user in the correct order.  For example, if the user asks for ingredient substitutions, the system should not respond by providing information based on the current step of the recipe. 

\subsection{Few-Shot Transfer Learning}
\label{subsec:fewshot_transfer}


Instruction-grounded chatbots could potentially be developed for many domains beyond cooking, for example, repair manuals \citep{wu-etal-2022-understanding}, wet-lab protocols \citep{kulkarni2018annotated}, software documentation \citep{olmo2021gpt3}, etc.  Each new domain will require a different set of user intents, motivating the need for few-shot intent classification. To better support few-shot learning, we also investigate whether existing large-scale dialogue corpora, such as MultiWOZ, can be used to transfer knowledge to instruction-grounded chatbots. 

Simply training on MultiWOZ using existing intent labels is unlikely to work well, because the MultiWOZ intents are very different from those needed for recipe-grounded dialogue.  To address this challenge, we utilize natural language descriptions of the intents, following \citet{zhao2022description}.  A full list of intent descriptions is provided as input to T5 model \cite{raffel2020exploring}, which is fine-tuned to predict the index of the correct intent.  Intent indices are randomized during fine-tuning, to prevent memorization of the MultiWOZ intents, which are different from those in ChattyChef, and force the model to learn to recognize intents based on their text descriptions.  A complete list of intents and their associated descriptions is presented in Table \ref{tab:intent_description} (in the Appendix). Example prompts used for MultiWOZ, and recipe-grounded conversation are presented in Table \ref{tab:dst_prompt} (in the Appendix).

In addition to supporting few-shot transfer learning from existing resources such as MultiWOZ, the intent descriptions are also useful for providing intent information to the response generation model (see \S \ref{sec:response_generation}, and Table \ref{tab:generation_prompt} in the Appendix for details.)

\begin{table}[t]
    \centering
    \small
    \begin{tabular}{cccc}
         \toprule
          & Train & Valid & Test \\
         \midrule
         \#Dialogues & 134 & 46 & 87 \\
          \#Turns per dialogue & 26.6 & 24.8 & 26.0 \\
         \bottomrule
    \end{tabular}
    \caption{Data split statistics of \ourdataset.}
    \label{tab:data_stats}
\end{table}

\subsection{Experiments}
\label{sec:experiments_intent}

\paragraph{Datasets.} To evaluate the performance of the model on \ourdataset~in the few-shot setting, we annotate the user intents for 10 conversations from the train set, 10 conversations from the validation set, and all 87 conversations in the test set. We consider a total of 19 different user intents for \ourdataset, which include 16 intents inherited from the CookDial dataset \cite{jiang2022cookdial} and 3 new intents: \texttt{req\_confirmation} (ask for verification), \texttt{req\_explanation} (ask to explain the reason or explain in more detail), \texttt{req\_description} (ask for description). The full list of all 19 intents and their descriptions can be found in Appendix \ref{sec:appendix_intent}. The numbers of intent annotations in the train, validation, and test split are 128, 125, and 1059. 
In addition to \ourdataset, for intent detection, we also use other datasets to conduct the cross-dataset experiments. In particular, we use one in-domain dataset CookDial and two large out-of-domain datasets, namely MultiWOZ 2.2 and Schema Guided Dialogue (SGD). Specifically, MultiWOZ 2.2 contains dialogues covering 8 domains (i.e., restaurant, hotel,
attraction, taxi, train, hospital, bus, and police); while SGD has conversations spanning over 20 domains (e.g., banks, events, media, calendar, travel) but also not include cooking. For MultiWOZ and SGD, we extract all user utterances, which have active intents along with the conversation histories. For CookDial, as there is no official data split, we split the data on the conversation level with the proportion of train, validation, and test set being 8:1:1. The sizes of the train/validation/test set of MultiWOZ, SGD and CookDial are 47,897/6,208/6,251, 153,940/22,832/39,623, and 3,599/466/545, respectively.

\paragraph{Models.} We choose to experiment with the following training settings to evaluate this subtask: (1) \textbf{In-context learning} (In-context): As a baseline approach, we prompt the GPT-J model, which learns to do this task by only observing a few examples without any parameter updates. (2) \textbf{Few-shot fine-tuning} (None \textrightarrow~\ourdataset): in this setting, we fine-tune the T5 model (following the approach discussed in \S \ref{subsec:fewshot_transfer}) on a few training examples (16-/128-shot) from \ourdataset. (3) \textbf{Cross-dataset} (X \textrightarrow~\ourdataset): the T5 model is first fine-tuned on another dataset (i.e., X may be MultiWOZ, SGD, or CookDial), and the fine-tuned model is then directly used to predict user intents in \ourdataset~ for 0-shot experiment or is further fine-tuned on a few examples of \ourdataset~to perform the task. (4) \textbf{Cross-dataset two-hop} (X \textrightarrow~CookDial \textrightarrow~\ourdataset): this setting is similar to the \textit{Cross-dataset} setting except that the model is fine-tuned on two datasets, first on an out-of-domain dataset (either MultiWOZ or SGD) then on CookDial.

For all models which utilize T5, we use the T5-XL version. More details about the training process of these models are described in Appendix~\ref{sec:appendix_intent}.


\paragraph{Results.} Following \citet{jiang2022cookdial}, we use micro-F1 as the evaluation metric for Intent Detection. Table~\ref{tab:intent_detection} demonstrates the performance of different models. In-context learning with 16 demonstrations significantly outperforms few-shot fine-tuning on a single dataset (None \textrightarrow~\ourdataset) with 128 examples.

Moreover, fine-tuning on another dataset first (X \textrightarrow~\ourdataset), either from in-domain or out-of-domain, does help boost the performance (over None \textrightarrow~\ourdataset) dramatically on all settings. This result is expected for the in-domain dataset as, besides the domain similarity, CookDial and \ourdataset~also share a large number of intents. More interestingly, leveraging MultiWOZ and SGD also improves the performance by more than 28\% and 33\% for the 16- and 128-shot, respectively, even though these two datasets cover quite different domains and intents from ChattyChef. 

Finally, fine-tuning the model on MultiWOZ or SGD first further improves the performance of CookDial \textrightarrow~\ourdataset. In particular, both MultiWOZ/SGD \textrightarrow~CookDial \textrightarrow~\ourdataset~outperform CookDial \textrightarrow~\ourdataset~on all settings with large margins. From this result and the above observations, we note that fine-tuning on a large dataset first, even from other domains, is extremely helpful for intent detection in the low-resource setting.

Since we want to measure the effectiveness of incorporating the intent information into the generation model, we will use the intent predictions from the best model (SGD \textrightarrow~CookDial \textrightarrow~\ourdataset~128-shot) in the later experiments on response generation (\S \ref{sec:response_generation}).

\section{Instruction State Tracking}
\label{sec:state_tracking}
We study the second subtask to support the instruction ordering of the generation module -- Instruction State Tracking. The goal of this task is to predict the current state of the instruction, or in other words,  the last instructed recipe step. Formally, given the $t^{\text{th}}$ system response $U^{sys}_t$, the previous instruction state $T_{t - 1}$ (i.e., an index of a recipe step), and the recipe with a list of $n_r$ steps $\mathcal{R} = \{R_1, R_2, \dots, R_{n_r}\}$, the expected output of this subtask is $T_{t}$.

\begin{table}[t]
\setlength{\tabcolsep}{2pt}
    \centering
    \setlength{\tabcolsep}{1pt}
    \resizebox{\linewidth}{!}{%
    \begin{tabular}{lccc}
        \toprule
        \multicolumn{1}{c}{\textbf{Model}} & \textbf{0-shot} & \textbf{16-shot} & \textbf{128-shot} \\
        \midrule
        In-context & - & 40.2 & - \\
        \midrule
        None\textrightarrow \ourdataset & - & 7.5 & 32.0 \\
        MultiWOZ\textrightarrow \ourdataset & 14.2 & 36.2 & 65.1 \\
        SGD\textrightarrow \ourdataset & 21.5 & 34.9 & 66.9 \\
        CookDial \textrightarrow \ourdataset & 72.3 & 72.8 & 74.5 \\
        \midrule
        MultiWOZ\textrightarrow CookDial\textrightarrow \ourdataset & \textbf{73.9} & \textbf{76.6} & 77.7 \\
        \midrule
        SGD\textrightarrow CookDial\textrightarrow \ourdataset & 73.7 & 76.5 &  \textbf{78.3} \\
        \bottomrule
    \end{tabular}}
    \caption{Performance of models in the User Intent Detection task in few-shot settings. Fine-tuning the models on large out-of-domain datasets first (i.e., MultiWOZ or SGD) is helpful for the task in low-resource settings.}
    \label{tab:intent_detection}
\end{table}

\begin{table}[t]
    \centering
    \small
    \begin{tabular}{cccc}
         \toprule
           & \# & WordMatch & SentEmb \\
          \midrule
          Validation & 576 & \textbf{82.0} & 80.8 \\
          Test & 1145 & 79.0 & \textbf{79.4} \\
         \bottomrule
    \end{tabular}
    \caption{The alignment accuracy for Instruction State Tracking on the validation and test set.}
    \label{tab:alignment_result}
\end{table}


\subsection{Aligning Conversations to Recipe Steps}
\label{state_tracking_approaches}
For this subtask, we adopt a simple unsupervised approach to track the instruction state. The key idea of our approach is to align the most recent system utterance with its most similar step in the recipe, and this aligned step will be the current instruction state. If the utterance can not be aligned with any recipe steps, the current instruction state will be the same as the previous one.  For the scoring function that measures similarity between the conversation history and the text of recipe steps, we use two simple approaches: (1) WordMatch (Word Matching): the scoring function computes the unigram F1 overlap between two input texts. (2) SentEmb (Sentence embedding): the scoring function computes the cosine similarity between sentence embeddings of the two input texts. More details about the alignment algorithm and SentEmb approach are described in Appendix~\ref{subsec:ist}.

\subsection{Experiments}

\paragraph{Setup.} For this subtask, we evaluate two approaches: \textbf{WordMatch} and \textbf{SentEmb}, which were discussed above. We manually annotate the instruction states for all the system responses in \ourdataset~and evaluate the accuracy of the two approaches on the validation and test sets. 

\paragraph{Results.} The performance of Instruction State Tracking is reported in Table \ref{tab:alignment_result}. Despite of its simplicity, the WordMatch approach has comparable performance to SentEmb. In particular, WordMatch outperforms SentEmb on the validation set by 1.2\%, but is slightly worse on the test set by 0.4\%. One plausible explanation is that there are many entities (e.g., ingredients, cooking utensils) in the recipe that are hardly be paraphrased in the cooking dialogue. In the next section, we will use the predicted instruction state from the WordMatch approach for integration with the generation model.

\section{Response Generation}
\label{sec:response_generation}
Given the conversation history and the grounded recipe, the Response Generation task aims to generate the instruction response to the user. Formally, given the history $\mathcal{H} = \{U^{sys}_1, U^{usr}_1, ..., U^{sys}_{t-1}, U^{usr}_{t-1}\}$, the recipe $\mathcal{R}$, the dialog system is expected to generate the next utterance ${U^{sys}_{t}}$.

\subsection{Generating Dialog Responses}

\paragraph{Base Model.}
\label{model}
In this work, we chose GPT-J \cite{gpt-j} as the base model. To fine-tune the model, we follow the approach of \citet{peng2022godel} that concatenates the dialog history, the cooking recipe, and the system response as ``$\mathcal{H} \texttt{ <|Knowledge|> } \mathcal{R} \texttt{ =>[system]}$ $U^{sys}_{t}$'' and feed it to the model. Both \texttt{[system]} and \texttt{<|Knowledge|>} are regular text strings. Let $S$ be the source text, which corresponds to part of this concatenated string of the dialog history and the cooking recipe  (i.e., ``\texttt{$\mathcal{H} \text{ <|Knowledge|> } \mathcal{R} \text{ =>[system]}$}''). In the fine-tuning phase, the model tries to learn the conditional probability of $P(U^{sys}_{t}|S)$, which can be written as the product of a series of conditional probabilities:
\begin{equation*}
    P(U^{sys}_{t}|S) = \prod_{i=1}^{n_t} p(U^{sys}_{t, i} | U^{sys}_{t, < i}, S)
\end{equation*}

\noindent where $n_t$ is the length of the response at the $t^{\text{th}}$ turn of conversation, $U^{sys}_{t, < i}$ indicates all tokens before the $i^{\text{th}}$ token generated by the system.



\paragraph{Intent-aware Model.} Since the intent labels may not convey the full meaning of the user's intents (\S \ref{sec:methods_intent}), we propose to leverage the natural language description of the intents  when integrating this information type into the model. In particular, we enhance the input prompt to the GPT-J model as ``$\mathcal{H} \texttt{ <|Knowledge|> } \mathcal{R} \texttt{ [user] wants}$ $ \texttt{to:} D. \texttt{ => [system] } U^{sys}_{t}$'', where $D$ is the description of user intent. An example prompt is shown in Table \ref{tab:generation_prompt} (in the Appendix).

\begin{table*}[htbp]
    \centering
    \small
    \setlength{\tabcolsep}{2pt}
    \begin{tabular}{lcccR{1cm}lc}
         \toprule
         \multicolumn{1}{c}{\multirow{2}{*}{\textbf{Model}}} &
         \multicolumn{5}{c}{\textbf{Automatic Evaluation}} & \textbf{Human Evaluation} \\ 
         \cmidrule(lr){2-6}  \cmidrule(lr){7-7} 
         & \textbf{BLEU} & \textbf{BLEURT} & \textbf{Length} & \multicolumn{2}{c}{\textbf{Diversity}} & {\color{myred} wrong order} | {\color{myorange} irrelevant} | {\color{myblue} lack of info.} | {\color{myblue2} wrong info.} | {\color{mygreen} correct}  \\ 
         \midrule
         GPT-J & 4.1 & 44.7 & 11.1 & 9.9\enspace/ & 37.9 & \raisebox{-0.22\height}{\includegraphics[scale=0.527]{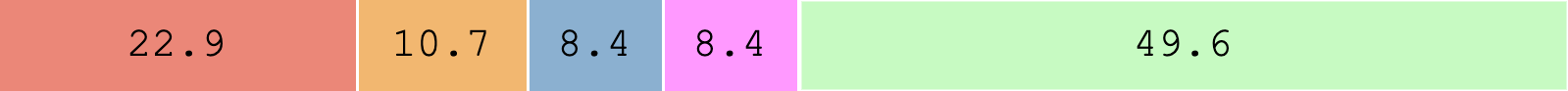}} \\
         GPT-J+int & 3.9 & 45.0 & 10.0 & 10.4\enspace/ & 38.5 & \raisebox{-0.22\height}{\includegraphics[scale=0.527]{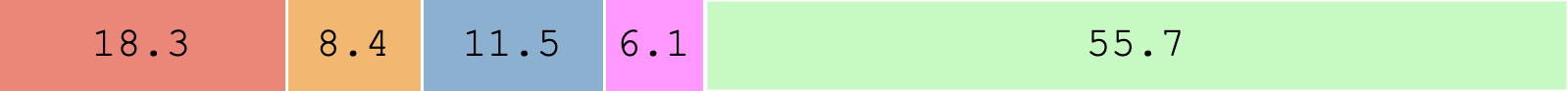}}\\
         GPT-J+cut & 4.3 & 45.2 & 10.9 & 9.9\enspace/ & 38.7 & \raisebox{-0.22\height}{\includegraphics[scale=0.527]{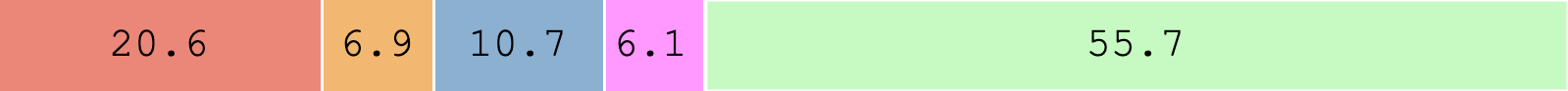}} \\
         GPT-J+ctr & \textbf{4.7} & \textbf{45.9} & \textbf{11.7} & 9.3\enspace/ & 36.6 & \raisebox{-0.22\height}{\includegraphics[scale=0.527]{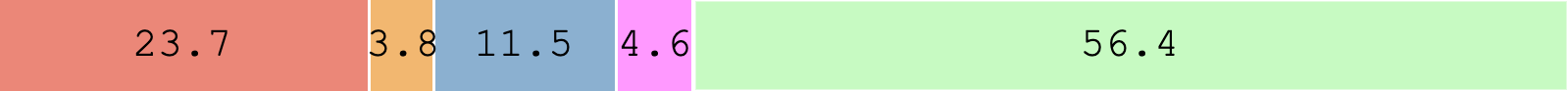}} \\
         GPT-J+ctr+int & 4.2 & 45.1 & 10.3 & \textbf{10.8}\enspace/ & \textbf{39.3} & \raisebox{-0.22\height}{\includegraphics[scale=0.527]{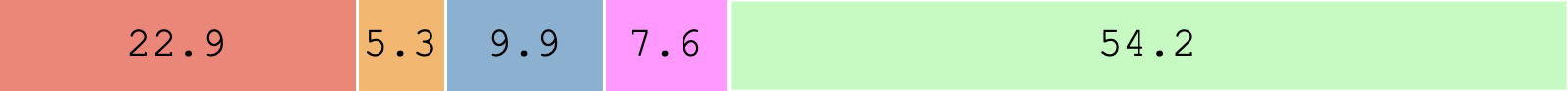}} \\ 
         \midrule
         ChatGPT\textsuperscript{$\dagger$} & 5.4 & 53.0 & 64.9 & 12.5\enspace/ & 45.3 & \raisebox{-0.22\height}{\includegraphics[scale=0.527]{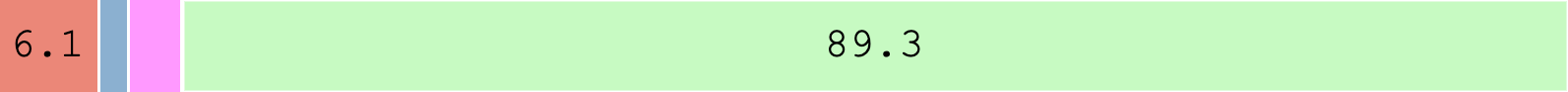}} \\
         \bottomrule
    \end{tabular}
    \caption{Automatic and human evaluation of Response Generation models \ourdataset. The automatic evaluation is conducted on the entire test set of 87 multi-turn conversations, and human evaluation is on 10 multi-turn conversations (131 generated responses). $\dagger$: Automatic evaluation metrics for ChatGPT are computed using the same subset of test data used in the human evaluation due to its access constraints. Length: the average length in terms of number of tokens. Diversity: diversity scores based on the unique unigrams (left) and bigrams (right). 
 }
    \label{tab:incorporate_result}
\end{table*}

\paragraph{State-aware Model.}
\label{state_aware}
When provided with a full recipe, the response generation model might have difficulty choosing the correct recipe part to condition on when generating a response, which can lead to giving wrong order instructions. As the instruction state (\S \ref{sec:state_tracking}) indicates the last instructed step, this information is essential for selecting the proper knowledge from the recipe for the model. Therefore, we explore two heuristic approaches for knowledge selection: (1) {\bf Cutoff:} only select recipe steps starting from the current instruction state. Formally, the input prompt to the GPT-J model is ``$\mathcal{H} \texttt{ <|Knowledge|> } \mathcal{R}' \texttt{=>[system]} U^{sys}_{t}$'', where $\mathcal{R}' = \{R_{T_{t-1}}, \dots , R_{n_r}\}$, and $T_{t-1}$ is the output from the Instruction State Tracking module. (2) {\bf Center:} Only select recipe steps in the $\pm$1 window around the current state. Formally, the input prompt to GPT-J will be similar to Cutoff except for $\mathcal{R}' = \{R_{T_{t-1} - 1}, R_{T_{t-1}}, R_{T_{t-1} + 1}\}$.

\subsection{Experimental Setup}
For this task, we evaluate the following models: (1) \textbf{GPT-J}: the base GPT-J model. (2) \textbf{GPT-J+cut}: a state-aware model, using the \textit{Cutoff} approach to select the grounded knowledge. (3) \textbf{GPT-J+ctr}: same as the above method, but the grounded knowledge is selected by using the \textit{Center} approach. (4) \textbf{GPT-J+int}: the GPT-J model which is incorporated with User Intent information. (5) \textbf{GPT-J+ctr+int}: a state-aware model using the \textit{Center} approach and is additionally incorporated with \textit{user intent}. (6) \textbf{ChatGPT}:\footnote{https://chat.openai.com/chat} We also interact with ChatGPT, a chatbot launched by OpenAI. We provide the recipes and the corresponding conversation histories from our test set and ask ChatGPT to provide the next system response. At the time of writing, because OpenAI had not yet published an API to process a large number of requests, we manually interacted with ChatGPT and collected 131 responses for 10 test conversations. The details about the training process of GPT-J base model and its variants are provided in Appendix \ref{sec:appendix_rg}.


\subsection{Results}

We report the following automatic evaluation metrics: BLEU \cite{papineni-etal-2002-bleu},\footnote{All BLEU scores reported in this paper are based on corpus-level BLEU-4.} BLEURT \cite{sellam-etal-2020-bleurt}, the average length of the outputs and the diversity scores \citep{li2016deep} based on the proportion of unique n-grams over the total number of n-grams.  Because there is a lack of consensus on how best to automatically evaluate open-domain dialogue models  \citep{liu2016not,sedoc2019chateval,csaky2019improving}, we also conduct two human evaluations in which model outputs are rated in terms of \textit{correctness} while the \textit{errors} are categorized.

\paragraph{Automatic Evaluation.}
Table~\ref{tab:incorporate_result} shows the performance of different models on the test set of \ourdataset. All GPT-J variants (+int, +cut, +ctr, and +int+ctr) have comparable or better performance than the base model, except for GPT-J+int, which has a lower BLEU score, and GPT-J+ctr, which has lower diversity scores. In terms of the comparison of the two knowledge selection methods, GPT-J+ctr has higher BLEU and BLEURT scores, while GPT+J+cut has better diversity scores. One possible reason is that the \textit{Center} approach considers a small context window of at most three recipe steps. It helps the model to focus on the most relevant information; but at the same time, it reduces the total amount of knowledge the model can rely on to instruct, making the responses less diverse than other models. Finally, incorporating the intent information does not show improvement in terms of BLEU and BLEURT, however, this approach does help increase the diversity of the generated responses. We also conduct additional experiments on the CookDial dataset in Appendix \ref{sec:cookdial_rg}.


\begin{figure*}[tb]
    \centering
    \includegraphics[width=\textwidth]{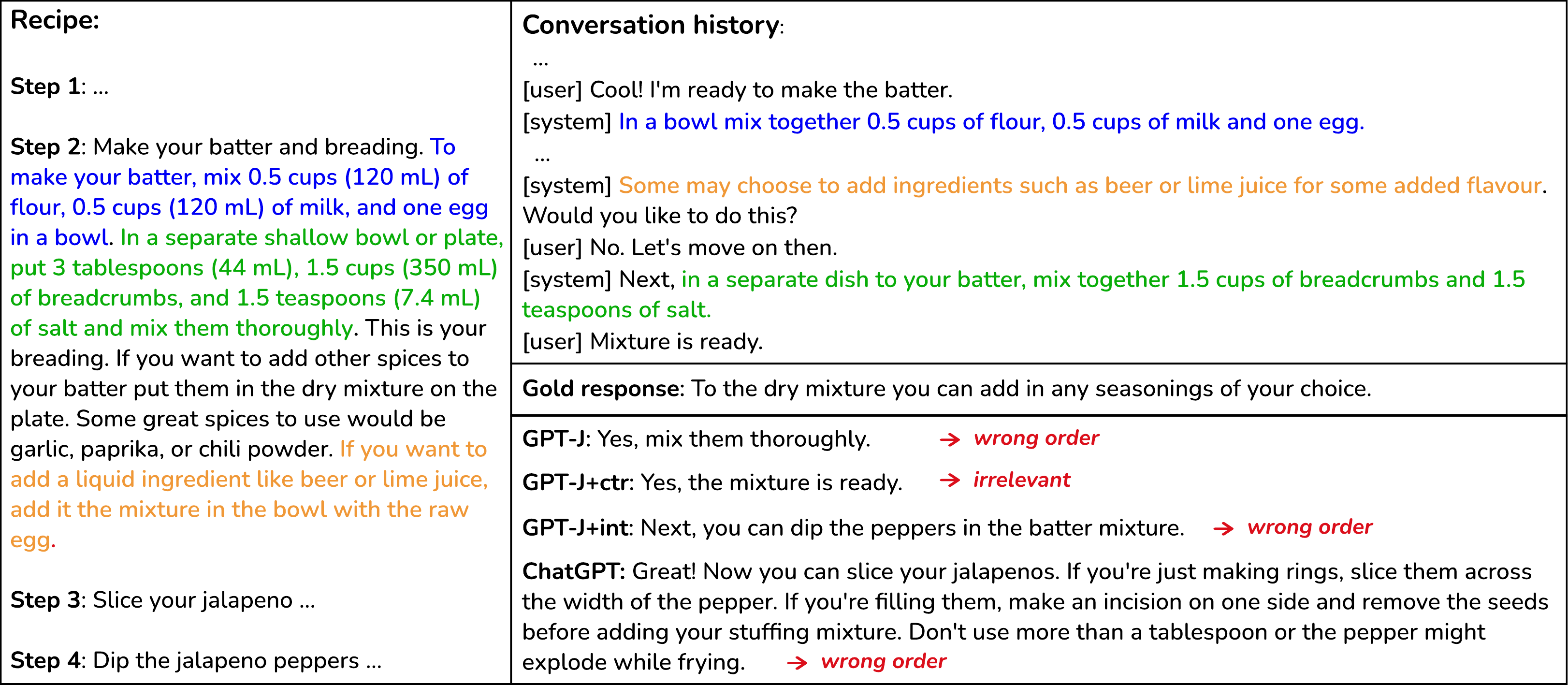}
    \caption{Example errors generated by each of the four systems. The system utterance in the history and its corresponding recipe micro-step have the same color. Text in \textcolor{red}{red} indicates the error types of the system responses.}
    \label{fig:example1}
\end{figure*}

\begin{figure}[tb]
    \centering    \includegraphics[width=0.98\columnwidth]{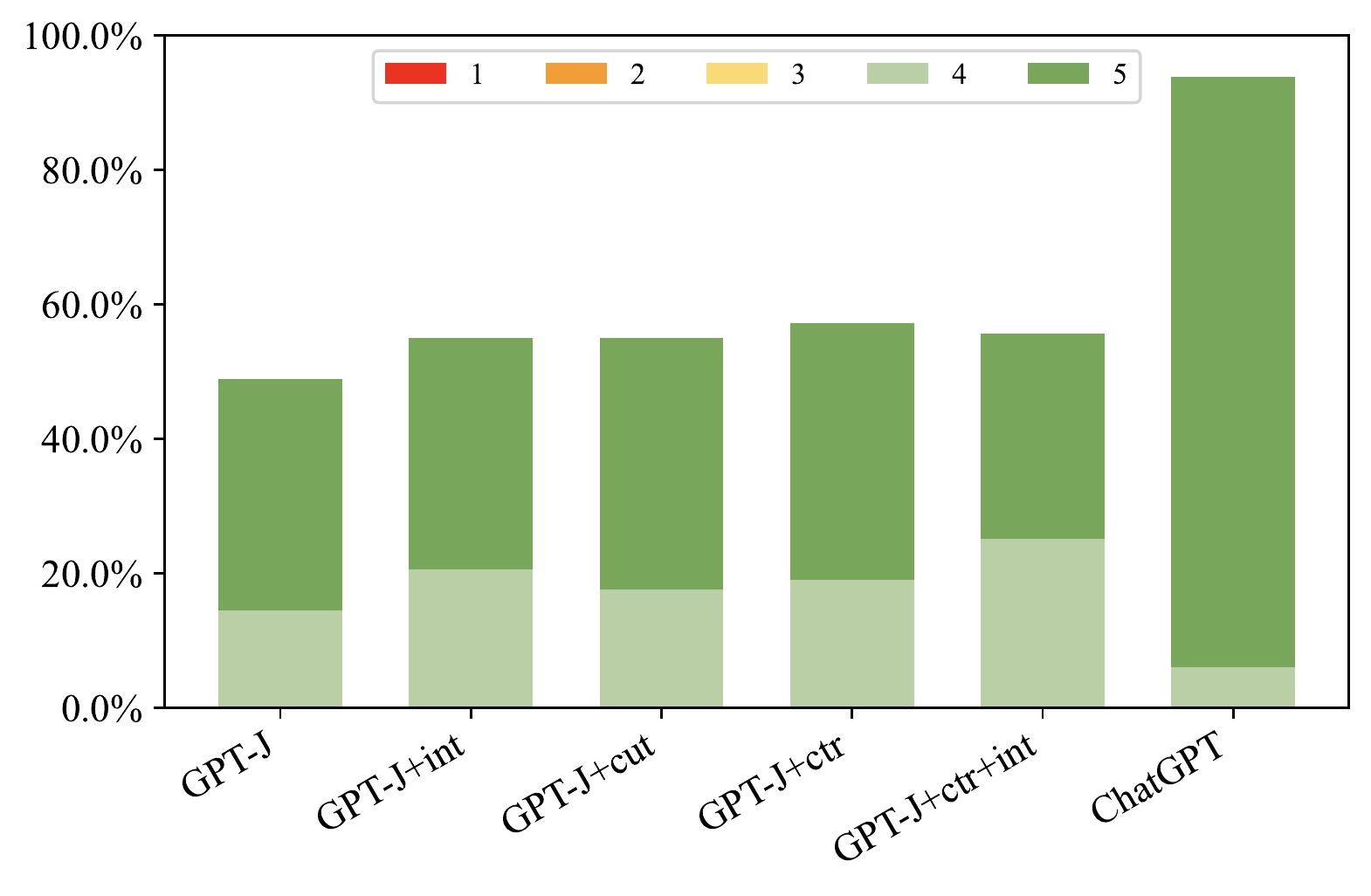}
    \caption{Percentages of model's outputs with high correctness ratings of 4 and 5. All GPT-J (+int/+cut/+ctr) variants that use intent and/or state tracking information generate better outputs  than the vanilla GPT-J.}
    \label{fig:correct_degree}
\end{figure}

\begin{figure}[tb]
    \centering
    \includegraphics[width=0.98\columnwidth]{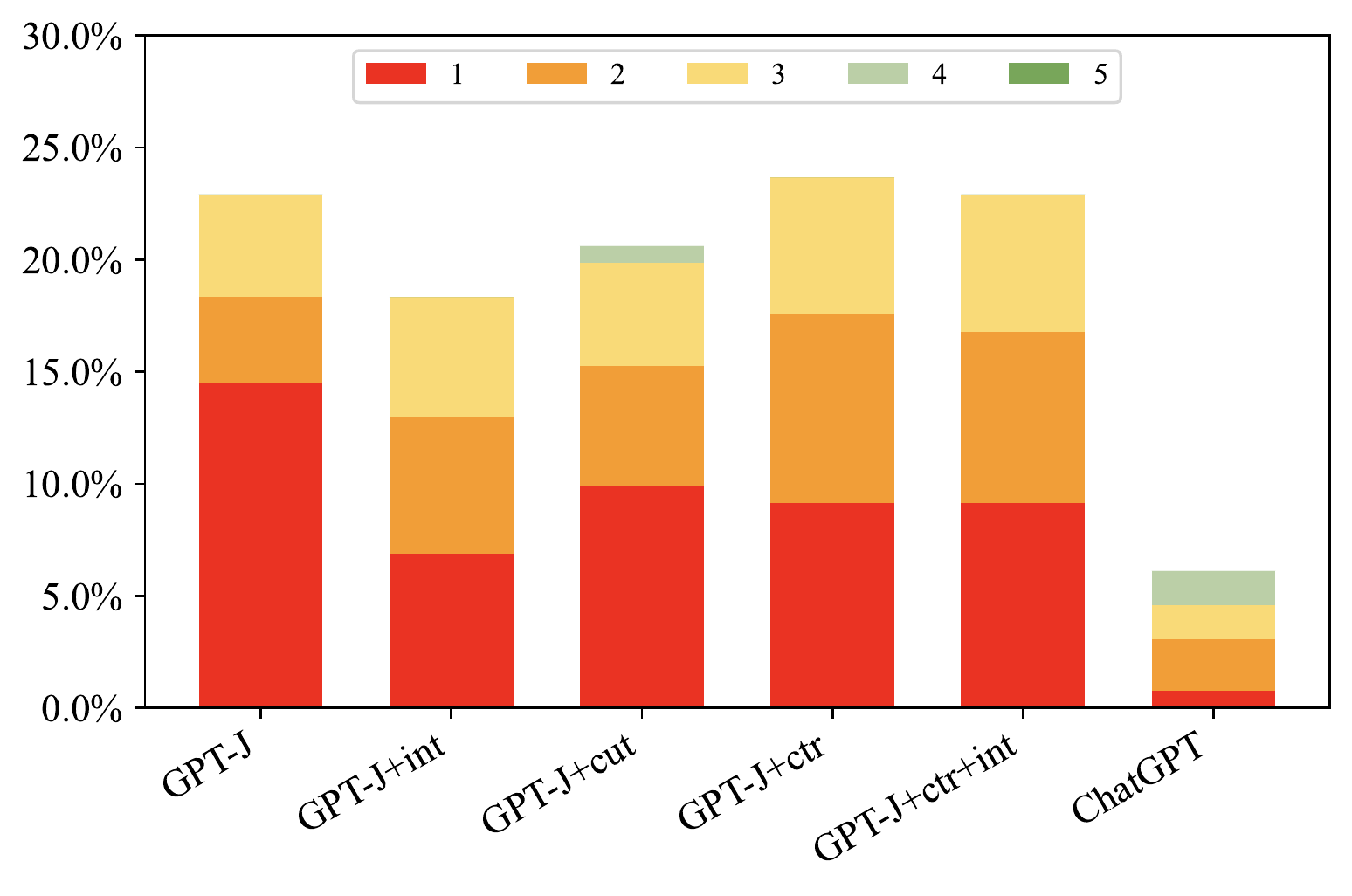}
    \caption{Correctness ratings of the wrong-order outputs from each model. All GPT-J (+int/+cut/+ctr) variants make fewer serious wrong-order errors (correctness of 1) than the vanilla GPT-J. Errors are sometimes not seen as severe in ChatGPT's longer responses than GPT-J's.}
    \label{fig:wrong_order_degree}
\end{figure}

\paragraph{Human Evaluation.}
In order to further understand the behaviors of the models, we ask three annotators to analyze the system outputs and manually categorize their errors. We also ask the annotators to rate the \textit{correctness} of each system response, using a 5-point Likert scale (i.e., 5-completely correct; 4-mostly correct, has minor and acceptable errors; 3-borderline, has both correct and incorrect information, nothing outweighs the other; 2-mostly incorrect, but still has correct information; 1-completely incorrect). The inter-annotator agreements measured by the nominal and ordinal Krippendorff’s alpha \cite{krippendorff2004reliability} for the error categorization and correctness rating are 0.43 and 0.62, respectively. More details about how to aggregate three annotations are in Appendix \ref{sec:human_eval}). As shown in Table~\ref{tab:incorporate_result} and Figure~\ref{fig:correct_degree}, all GPT-J variants that incorporate the intent and/or state information have fewer errors and more responses rated with 4 and 5 for correctness than the base model. Even though automatic metrics do not show a clear difference, human evaluation reveals that GPT-J+int has fewer (22.9\%$\rightarrow$18.3\%) wrong-order errors compared to the base model  and is also the model with the least number of this type of error. On the contrary, using \textit{Center} approach (i.e., GPT-J+ctr and GPT-J+ctr+int) in grounded recipe selection does not have much impact on reducing the number of wrong-order responses, despite the fact that it helps improve BLEU and BLEURT scores. In addition,
all +int/+ctr variants of GPT-J have fewer responses with severe wrong-order errors (\textit{correctness} of 1) than the base model.

Finally, we also analyze the errors in the outputs of ChatGPT. Overall, ChatGPT performs extremely well in this task with only 10.7\% of the outputs being erroneous. The outputs of ChatGPT are notably longer than other systems since ChatGPT tends to instruct multiple recipe steps in one utterance or utilize knowledge outside the given recipe. As shown in Table \ref{tab:incorporate_result}, wrong-order instruction is still the most common error for ChatGPT. One scenario where ChatGPT makes mistakes in terms of the ordering is when the recipe step contains multiple micro-steps (see an example in Figure~\ref{fig:example1}). It indicates that there are still many challenges that remain unsolved in the cooking instruction dialogue task.
\section{Related Work}
The task of recipe-grounded conversation is close to the Conversational Question Answering (CQA) task. In CQA, given a reference text, the system needs to engage in a multi-turn conversation to answer questions from users. Compared to single-turn question answering, CQA raises new challenges (e.g., co-reference resolution, contextual reasoning) due to the dependency between Question Answering turns. There exist multiple datasets in this area, such as CoQA \cite{reddy2019coqa}, QuAC \cite{choi-etal-2018-quac}, DoQA \cite{campos-etal-2020-doqa}, and ShARC \cite{saeidi2018interpretation}. There are several differeces between Instructional Dialogue and Conversational Question Answering. Firstly, in the dialogue setting, the message from the system can also be a question, such as verification. Secondly, while the goal of CQA is seeking information, Instructional Dialogue focuses on supporting users to complete a procedure; therefore, there is additional order-related relationship between the system's responses and the instructions that needs to be managed by the dialog agent.

Recent work has investigated issues that arise in chatbots based on large language models.  For instance, they are known to sometimes generate toxic language \citep{baheti2021just,deng2022cold}, make factual errors in their statements \citep{honovich2021q2,dziri2022faithdial,rashkin2021increasing}, and be overly confident \citep{mielke2022reducing}.  In this work, we focus on addressing a specific problem related to instruction-grounded dialogue, which is presenting information in the wrong order to a user.

A small amount of prior work \cite{jiang2022cookdial, strathearn-gkatzia-2022-task2dial} has started to explore the problem of recipe-grounded conversation, which makes these papers the two closest to ours.
 Both of these papers focused primarily on dataset creation.  \citet{jiang2022cookdial} included experiments on response generation, but as their focus was on building a new dataset, they did not conduct extensive experiments or perform a human evaluation of their system's outputs.  They did propose baselines and evaluate the tasks of User Question Understanding and Agent Action Frame Prediction, which are similar to our User Intent Detection and Instruction State Tracking. Although these tasks have similar goals, our work is different in the sense that we focus on tackling the problems in the low-resource setting, by transferring knowledge from existing dialogue corpora such as MultiWOZ. Finally, besides providing additional recipe-grounded conversations as in these two prior works, our main focuses are on analyzing challenges of current large langauge models (i.e., GPT-J and ChatGPT) on this task and addressing the specific challenge of instruction ordering.

\section{Conclusion}

In this paper, we have proposed to explore two additional subtasks, namely User Intent Detection and Instructional State Tracking, to mitigate the problem of incorrect instruction order in Instructional Dialogue. We analyze these two auxiliary subtasks with different methods in low-resource settings. Even though the performance of the modules for the two subtasks is still low, experiment results show that incorporating the user intent or state information does help to mitigate the wrong-order instructions issue, with the intent information having a greater impact. However, combining the two pieces of information does not lead to improvement over using each individual one of them alone. Therefore, we believe that further research for the two subtasks is still needed, and also more effective ways of incorporating the information into the Response Generation module need to be investigated. Finally, we release \includegraphics[width=1em]{images/cook_1f9d1-200d-1f373.png} \ourdataset, a new cooking instructional dataset, to promote future research in this direction.

\section*{Limitations}
In this work, we have only analyzed the common errors of two models (i.e., GPT-J and ChatGPT) in the Instructional Dialogue task. One open question is whether other GPT-based models or models with other architectures (e.g., encoder-decoder models) also have the same issue in this task. Our work and dataset are also limited to the English language. 

\section*{Ethical Considerations}
To collect recipe-grounded conversations we hired crowd workers using the Prolific platform.\footnote{\url{https://www.prolific.co/}}  The study was conducted with the approval of our local IRB. The compensation was derived based on Prolific's payment principles. We estimate the hourly pay for crowd workers was \$15.49 (details in Appendix \ref{sec:dataset_cost}). Crowd workers were strictly asked not to write any offensive content or personal information. 

\section*{Acknowledgments}
We thank Yao Dou, Fan Bai as well as four anonymous reviewers for their helpful feedback on this work. We also thank Govind Ramesh, Grace Kim, Yimeng Jiang for their help with human evaluation. This research is supported in part by the NSF awards IIS-2112633 and IIS-2052498. The views and conclusions contained herein are those of the authors and should not be interpreted as necessarily representing the official policies, either expressed or implied, of NSF or the U.S. Government. The U.S. Government is authorized to reproduce and distribute reprints for governmental purposes notwithstanding any copyright annotation therein.

\bibliography{custom}
\bibliographystyle{acl_natbib}
\clearpage
\appendix

\begin{figure*}[b]
    \centering
    \includegraphics[width=0.95\linewidth]{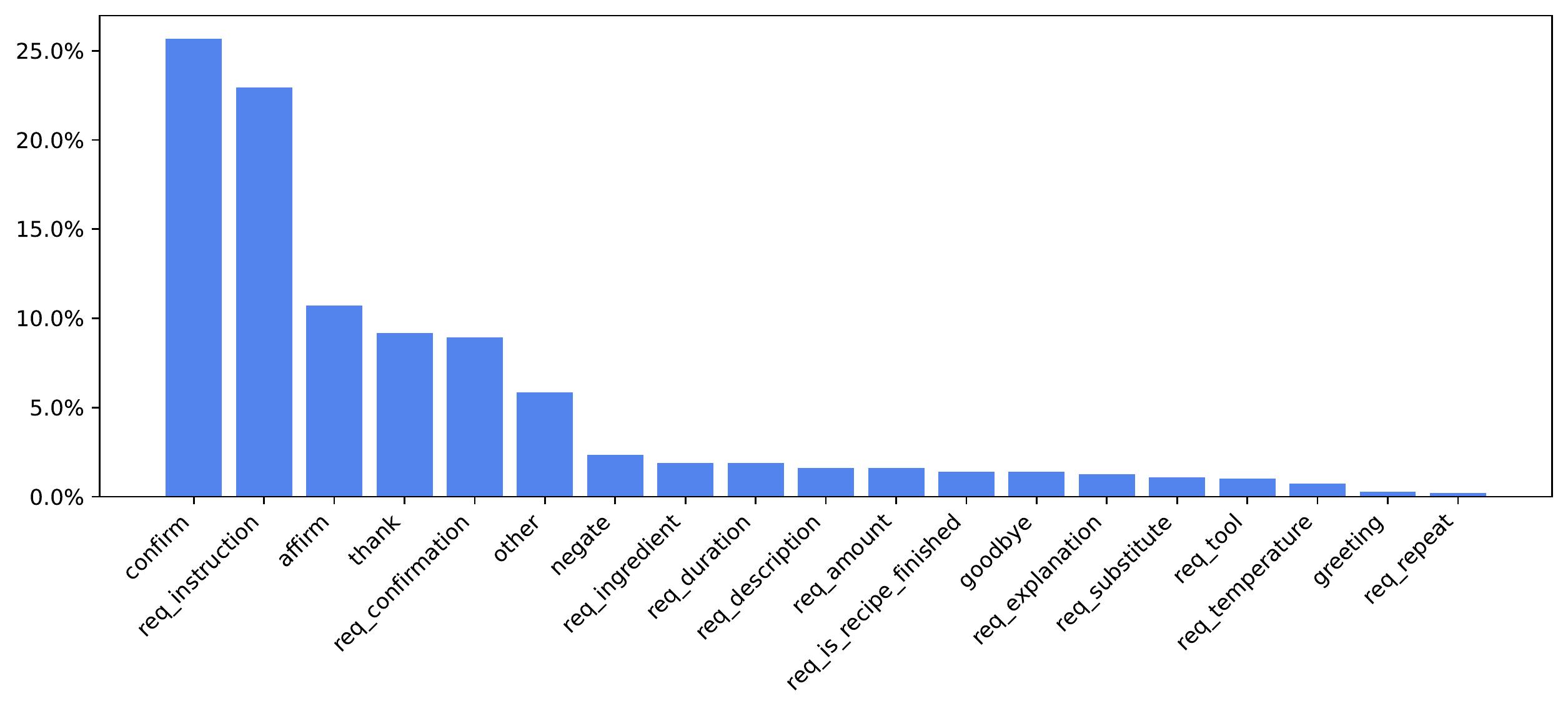}
    \caption{Statistic of user intents over 91 conversations.}
    \label{fig:intent_stats}
\end{figure*}

\section{Experiments on CookDial dataset}
\label{sec:cookdial_rg}
\subsection{Incorporation of Instruction State and User Intent information}
In this section, we explore the performance of models using Instruction State and Intent Information on the CookDial dataset. We fine-tuned the models adopting the same approaches as in \S \ref{sec:response_generation} and using the same set of automatic metrics to evaluate. Instead of extracting silver labels, We use the gold Instruction State (i.e., "tracker\_completed\_step" in CookDial) and User Intent information from the CookDial dataset. The performance of all models is demonstrated in Table \ref{tab:cookdial_results}. 

From the table, we can see that the model performance on CookDial is much higher than in our dataset, this is due to the more straightforward instruction scenarios and higher lexical similarity between the grounded recipe and system utterances in CookDial (as discussed in \S \ref{subsec:dataset_stats}). In addition, similar behaviors of the models fine-tuned on CookDial and \ourdataset~can also be observed here - incorporating the Instruction State information helps to improve BLEU and BLEURT scores, while incorporating the intent information helps the models generate more diverse responses.

\begin{table}[b]
    \centering
    \small
    \setlength{\tabcolsep}{2pt}
    \begin{tabular}{lcccR{1cm}l}
         \toprule
          \textbf{Model} & \textbf{BLEU} & \textbf{BLEURT} & \textbf{Length} & \multicolumn{2}{c}{\textbf{Diversity}}\\
         \midrule
         GPT-J & 34.7 & 65.0 & 11.5 & 13.0\enspace/ & 44.3  \\
         GPT-J+int & 30.9 & 62.5 & 10.4 & \textbf{13.8}\enspace/ & \textbf{45.3} \\
         GPT-J+cut & \textbf{35.6} & 65.2 & \textbf{11.7} & 12.8\enspace/ & 43.9 \\
         GPT-J+ctr & 35.3 & \textbf{65.4} & 11.3 & 13.3\enspace/ & 44.6  \\
         GPT-J+ctr+int & 32.0 & 63.2 & 10.6 & 13.7\enspace/ & 44.5 \\
         \bottomrule
    \end{tabular}
    \caption{Performance on the test set of CookDial dataset. }
    \label{tab:cookdial_results}
\end{table}

\subsection{Transfer learning from CookDial to \ourdataset}
To explore whether the information learned from CookDial is helpful for our dataset, we keep finetuning the model from Table \ref{tab:cookdial_results} with the corresponding settings on our dataset. As shown in Table \ref{tab:transfer_cookdial}, transfer learning does not show a clear improvement in terms of BLEU and BLEURT, it even hurts the performance of the model in many cases. The big difference between the two datasets (as discussed in \S \ref{subsec:dataset_stats}) may be the reason why transfer learning is ineffective here. However, transfer learning also has the merit of making the system outputs more diverse, especially for GPT-J+cut and GPT-J+ctr.

\begin{table}[tb]
    \centering
    \small
    \setlength{\tabcolsep}{2pt}
    \begin{tabular}{lccccc}
        \toprule
        \textbf{Model} & \textbf{BLEU} & \textbf{BLEURT} & \textbf{Length} & \multicolumn{2}{c}{\textbf{Diversity}}\\
         \midrule
        \multirow{2}{*}{GPT-J} & 3.7 & 45.3 & 9.7 & 11.0 & 39.9 \\
         & (-0.4) & (+0.6) & (-1.4) & (+1.1) & (+2.0) \\
         \midrule
        \multirow{2}{*}{GPT-J+int} & 4.3 & 45.5 & 10.4 & 11.2 & 40.8 \\
         & (+0.4) & (+0.5) & 0.4 & (+0.8) & (+2.3) \\
         \midrule
        \multirow{2}{*}{GPT-J+cut} & 4.0 & 45.2 & 9.9 & 11.8 & 44.6 \\
         & (-0.3) & (0.0) & (-1.0) & (+1.9) & (+5.9) \\
         \midrule
        \multirow{2}{*}{GPT-J+ctr} & 4.1 & 45.5 & 9.6 & 11.6 & 42.4 \\
         & (-0.6) & (-0.4) & (-2.1) & (+2.3) & (+5.8) \\
         \midrule
        \multirow{2}{*}{GPT-J+ctr+int} & 4.2 & 45.7 & 9.8 & 11.5 & 41.9 \\
         & (0.0) & (+0.6)  & (-0.5) & (+0.7) & (+2.6) \\
         \bottomrule
    \end{tabular}
    \caption{Transfer learning performance of models on the test set of \ourdataset. The number below in each cell indicates the change to the model fine-tuned with the same setting on \ourdataset~only.}
    \label{tab:transfer_cookdial}
\end{table}

\section{Implementation Details}

For all experiments, we train models across 4 A40 GPUs (48GB each). The total GPU hours for training a GPT-J model in the Response Generation task is about 5.3 hours, and the total GPU hours for training the T5 models in the User Intent Detection task is about 4 hours.

\subsection{User Intent Detection}
\label{sec:appendix_intent}
The details about user intents and their description are reported in Table~\ref{tab:intent_description}. Examples of input and out prompts to the T5 model are demonstrated in Table~\ref{tab:dst_prompt}.

For all experiments which use T5 model, we set the maximum sequence length to 1028 and the number of training epochs to 30, we stop the training process if the perplexity of the model on the validation set does not improve after 5 epochs. We use AdamW to optimize and consider the initial learning rate $\in$ \{1e-5, 5e-5, 1e-4\}. For the in-context setting, the maximum length of the input sequence is set to 1984.

For all the models, we employ beam search with a beam size of 5 for decoding. We select the model checkpoint that produces the lowest perplexity on the validation set and then apply the selected one to the test set.

\begin{table}[t!]
    \centering
    \small
    \begin{tabular}{p{0.35\linewidth}|p{0.55\linewidth}}
         \toprule
         Intents & Descriptions \\
         \midrule
          greeting & greeting \\  
          req\_temperature & ask about the cooking temperature  \\
          thank & thank \\
          req\_instruction & ask for instructions \\
          confirm & confirm the current stage \\
          req\_repeat & ask to repeat the last information \\
          negate & negate \\
          req\_amount & ask about the amount information \\
          req\_ingredient & ask about the ingredients \\
          req\_is\_recipe\_finished & ask whether the recipe is finished \\
          req\_tool & ask about the cooking tool \\ 
          req\_duration & ask about the cooking duration \\
          affirm & affirm \\
          goodbye & goodbye \\
          \fakemultirow{req\_substitute} & ask for tool or ingredient substitutions \\
          req\_confirmation & ask for verification \\
          req\_description & ask for the description \\
          \fakemultirow{req\_explanation} & ask to explain the reason or explain in more detail \\
          other & other intent \\
          \bottomrule
    \end{tabular}
    \caption{Descriptions of user intents in \ourdataset}
    \label{tab:intent_description}
\end{table}

\begin{algorithm}[h!]
\caption{Instruction State Tracking}
\small
\label{alignment}

\KwInput{
    Recipe $\mathcal{R} = \{R_1, R_2, \dots , R_{n_r}\}$ \newline
    Most recent system utterance: $U_t^{sys}$ \newline
    Previous instruction state: $T_{t-1}$
    Threshold parameters $0 < \alpha_1 \leq \alpha_2 < 1$ \newline
    Scoring function $f$
}

\KwOutput{Instruction State $T_t$}

Initialize $score[i] = 0 $ $\forall i = 1, 2, \dots , n_r$

Initialize $current\_state = T_{t-1}$

\tcc{compute similarity scores between the most recent system utterance and all recipe steps}
\For{$i = 1, 2, \dots , n_r$}{
    \label{first_for_line}
    $micro\_steps \leftarrow$ sentence\_tokenize($R_i$)
    
    $score[i] = \max_{r \in micro\_steps} f(U_t^{sys}, r)$
}

$best\_state \leftarrow \argmax (score)$

$max\_score \leftarrow score[best\_state]$
\label{max_score_line}

\If{
($best\_state == current\_state + 1$ and $max\_score > \alpha_1$) or ($max\_score > \alpha_2$)
}{
    $current\_state \leftarrow best\_state$
    \label{align_new_step}
}

$T_t \leftarrow current\_state$

\end{algorithm}

\subsection{Instruction State Tracking}
\label{subsec:ist}

The Instruction State Tracking Algorithm is described in Algorithm~\ref{alignment}). In order to produce the system utterance - recipe alignment, similarity scores between the most recent system response to all the recipe steps are computed (Line~\ref{first_for_line} - \ref{max_score_line} in Algorithm~\ref{alignment}). After that, by comparing the similarity score to thresholds (i.e.,  $\alpha_1$ and $\alpha_2$), the algorithm decides whether the current system utterance is aligned to a new state (i.e., a new recipe step -- line \ref{align_new_step} in Algorithm~\ref{alignment}) or is aligned with the previous state.

For the Sentence Embedding approach, we use \texttt{sentence-transformers/paraphrase\-MiniLM-L6-v2} \cite{reimers-gurevych-2019-sentence} to compute the sentence embeddings of system responses and recipe steps. We use NLTK\cite{bird2009natural} to perform the word and sentence tokenization. About the threshold, we set $\alpha_1, \alpha_2$ equal to 0.2 and 0.3, respectively, for the Word Matching approach. For the sentence embedding approach, we set $\alpha_1$ equal to 0.5 and $\alpha_2$ equal to 0.6. The thresholds are chosen based on the accuracy of the validation set.

\subsection{Response Generation}
\label{sec:appendix_rg}
An example of the input to the GPT-J model is illustrated in Table~\ref{tab:generation_prompt}.

To fine-tune all the GPT-J models, We set the maximum sequence length to 1280 and the number of training epochs to 3. We use AdamW to optimize and set the initial learning equal to 1e-5, except for transfer learning experiments with CookDial, in which we use the learning rate of 5e-6. We employ beam search with a beam size of 5 for decoding. We select the model checkpoint that produces the lowest perplexity on the validation set and then apply the selected one to the test set.

\begin{table}[b]
\small
\centering
\begin{tabular}{lcccc}
    \toprule
    & Paired & Self & M-User \\
    \midrule
    Avg time per turn (min) & 2.35       & 1.77  & 1.60 \\ 
    \midrule
    Avg cost per turn (\$) & 0.72 & 0.42 & 0.35 \\ 
    \midrule
    \#Turns per recipe step & 2.04 & 2.30 & 2.13 \\ 
    \midrule
    \#Dialogues  & 86  &  160 &  21 \\ 
    \bottomrule
    \end{tabular}
    \caption{Statistics of the data collection methods.}
    \label{tab:collection_stat}
\end{table}

\begin{table*}[tb]
    \small
    \centering
    \begin{tabular}{p{0.95\linewidth}}
         \toprule
         \textbf{Input:} \textcolor{red}{0:book a table at a restaurant 1:book a hotel to stay in 2:search for police station 3:search for places to wine and dine 4:search for a medical facility or a doctor 5:search for a bus 6:search for a hotel to stay in 7:search for trains that take you places 8:search for places to see for leisure 9:book taxis to travel between places 10:book train tickets} \textcolor{blue}{[user] I am looking for a hotel called the alpha-milton guest house. [system] Sure! I've located the guesthouse, it is located in the north area. Would you like me to book you a room? [user] No thank you but I do need the address please?} \\
         \\
         \textbf{Output:} [intents] 6 \\
         \midrule
         \textbf{Input}: \textcolor{red}{0:negate 1:confirm the current stage 2:ask to repeat the last information 3:ask about cooking duration 4:ask for verification 5:thank 6:ask to explain the reason or explain in more detail 7:ask about the cooking temperature 8:affirm 9:greeting 10:ask for the description 11:ask about the amount information 12:goodbye 13:ask whether the recipe is finished 14:ask for instructions 15:ask about the ingredients 16:ask about the cooking tool 17:ask for tool or ingredient substitutions 18:other intent} \textcolor{blue}{[user] Yes, what do I need? [system] Russet potatoes, or other high starch potatoes [user] What is the first step? [system] Wash and peel the potatoes, use cold water when washing [user] What do I use to peel the potatoes?}\\
         \\
         \textbf{Output}: [intents] 16 \\
         \bottomrule
    \end{tabular}
    \caption{Examples of the input and output of the User Intent Detection model. The top example is from MultiWOZ 2.2 dataset, and the bottom one is from Our cooking instruction dataset). \textcolor{red}{Red}: Indexed intent descriptions. \textcolor{blue}{Blue}: Conversation history.}
    \label{tab:dst_prompt}
\end{table*}

\begin{table*}[tb]
    \small
    \centering
    \begin{tabular}{p{0.95\linewidth}}
         \toprule
         \textcolor{blue}{[system] Would you like to learn how to make hash browns? [user] Yes, what do I need? [system] Russet potatoes, or other high starch potatoes [user] What is the first step? [system] Wash and peel the potatoes, use cold water when washing [user] What do I use to peel the potatoes?} <|Knowledge|> \textcolor{brown}{- Peel the potatoes. Wash the potatoes well in cold water, then peel using a small knife or a vegetable peeler. Russet potatoes, or other potatoes with a high starch content, work best for hash browns. - Shred the potatoes. Line a bowl with a clean dishtowel, then shred the potatoes directly into the towel-lined bowl, using a cheese grater. - Squeeze out the moisture. You must squeeze out as much moisture as possible from the shredded potatoes. This is the most important step in achieving crispy (rather than mushy) hash browns. To do this, gather the corners of the dishtowel containing the shredded potatoes and twist the neck until you form a tight package. Continue twisting the cloth and squishing the potato in your fist until you've squeezed as much liquid as you can from the potato. Alternatively, you can try squeezing the moisture from the potatoes using a potato ricer. You do not need to force the potatoes through the ricer, simply use it to press out the moisture. - Heat the skillet. Heat a large skillet pan (preferably cast iron) over a medium-high heat. Add the butter to the pan and allow to melt. Once the butter has melted, add the dry, shredded potatoes to the pan and toss to coat with butter. Season with salt and pepper. - Cook the hash browns. Once the potato has been coated with butter, flatten it using a spatula to maximize contact with the hot pan. It should be no more than 1/2 an inch thick. Cook for 3-4 minutes on the first side, flip, then cook for 2-3 minutes on the other side. The hash brown potatoes are ready when each side is crisp and golden brown. - Serve. Slide the hash brown from the pan, or lift using a large spatula. Cut it into halves or quarters, if necessary. Serve on its own, with hot sauce or ketchup, or alongside bacon and eggs for a top notch breakfast.} \textcolor{mygreen}{[user] want to: ask about the cooking tool.} => [system] \textcolor{red}{you can use a vegetable peeler, or a small knife} \\
         \bottomrule
    \end{tabular}
    \caption{An example of the prompt to the Response Generation model. \textcolor{blue}{Blue: Conversation history}. \textcolor{brown}{Brown}: Grounded recipe. \textcolor{mygreen}{Green}: Intent description prompt. \textcolor{red}{Red}: Output of the model.}
    \label{tab:generation_prompt}
\end{table*}

\section{Human Evaluation}
\label{sec:human_eval}
In this section, we discuss the way to aggregate annotations from annotators. For the error categorization experiment, the final decision of an example is reached if all three annotators have the same annotation. When only two annotators have the same annotation, a fourth one will join and decide whether to agree with the majority. In all other situations, a discussion between annotators is held, and the final decision is based on majority voting. For the correctness rating experiment, the rating of each example is the average score of the three annotators. In cases when only two annotators rate an example as completely correct or a rating of 5, but the third one detects an error and marks the example as incorrect (i.e., belongs to one of the four error types), if the final decision from the error categorization is also incorrect, the correctness of this example is the rating of the third annotator. The same rule applies to the opposite situation, i.e., only two annotators rate an example as completely incorrect, and the third one thinks it is correct.

\section{Dataset construction}
\label{sec:dataset_cost}
\subsection{Collection Strategies}
Even though employing two workers using different interfaces has advantages, we see that pairing workers on a task is inefficient. In particular, for each conversation, one worker would need to wait for a long time until the partner joined the task. Moreover, in some cases, some workers are uncooperative. For example, during the chat, one worker may spend too much time sending his messages or even quit the task, which will have a bad impact on his partner \cite{choi-etal-2018-quac, reddy2019coqa}.  
As a result, we study three different collection strategies as follows.

\noindent \textbf{Paired conversations} (Paired): Two workers are required for each conversation; one acts as the agent, and the other act as the user. After two workers got paired, they would be assigned to the same cooking task; however, only the agent has access to the recipe, and all the user knows about this task is the title (i.e., what to cook). 

\noindent \textbf{Self-chat} (Self): In this mode, only one worker is needed for each conversation. The worker will play both roles (i.e., Agent and User).

\noindent \textbf{Model-User} (M-User): One worker is assigned for each conversation in this mode. However, unlike Self-chat, when the worker plays the User role, he would be provided with candidate responses from a model, and he could either pick one and edit it or enter his own words.

All conversations were collected through the ParlAI API \cite{miller-etal-2017-parlai}. The participants for this paper were recruited using Prolific (\url{www.prolific.co}). We restricted the task to workers whose first language is English and with more than 100 submissions with at least a 99\% approval rate. Crowd workers were not informed of the specific details of the dataset. However, they consented to have their responses used for this purpose through the Prolific Participation Agreement.
The statistic about each collection method is reported in Table~\ref{tab:collection_stat}.

\subsection{Collection Interfaces}
\label{appendix:collection_interface}

\begin{figure*}[t]
\centering
\includegraphics[width=0.9\textwidth]{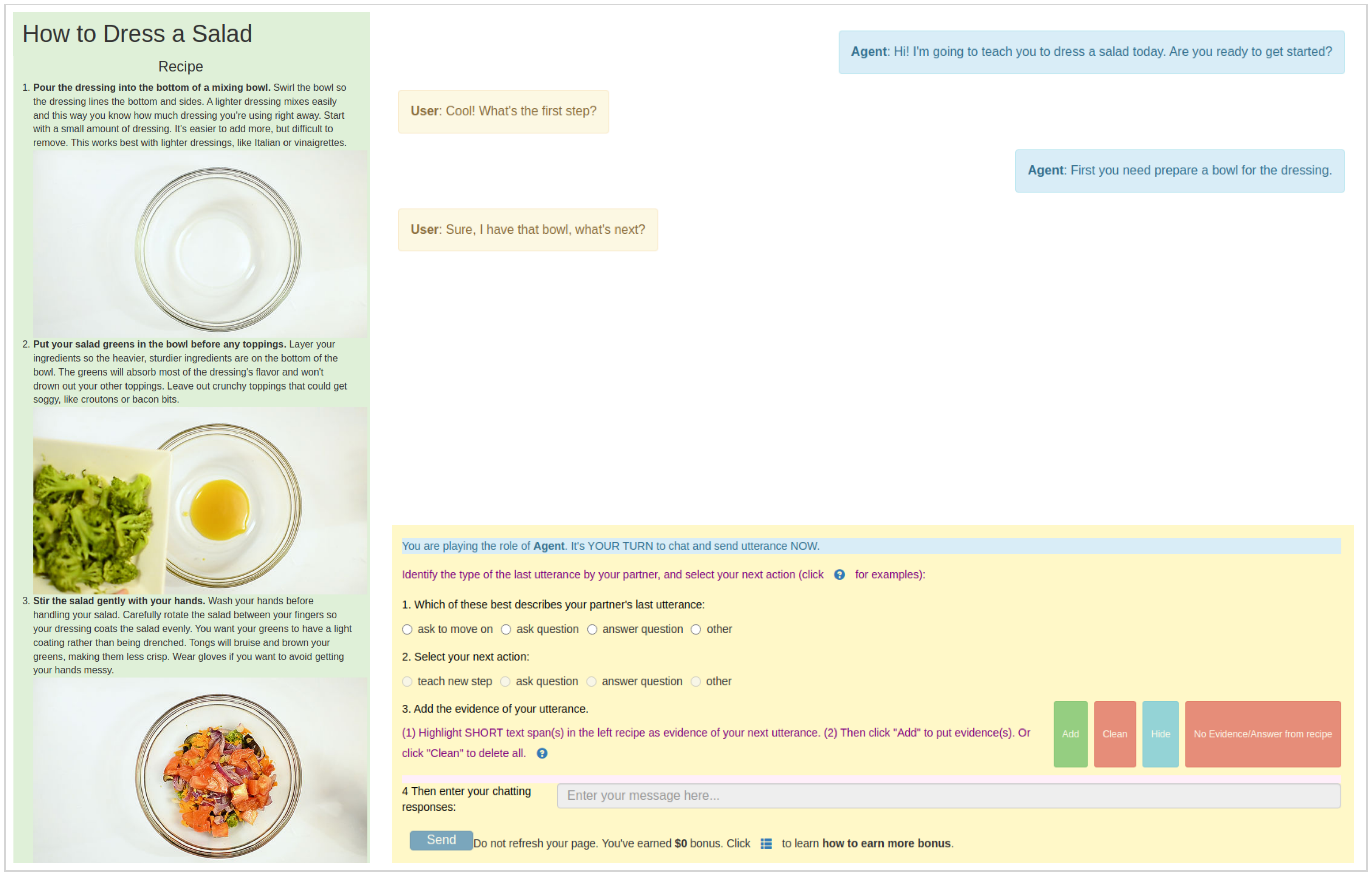}
\label{agent_interface}
\caption{A screenshot of our crowdsourcing interface.}
 \label{interface_examples}
\end{figure*}

See Figure \ref{interface_examples} for the screenshot of our crowdsourcing interface.

\end{document}